\documentclass[a4paper, 10pt, conference]{ieeeconf}      
\usepackage{FG2024}
\usepackage{times}
\usepackage{epsfig}
\usepackage{graphicx}
\usepackage{amsmath}
\usepackage{amssymb}
\usepackage{multirow}
\usepackage{xcolor}
\usepackage{booktabs}
\usepackage{gensymb}
\usepackage[pagebackref,breaklinks,colorlinks]{hyperref}
\FGfinalcopy 

\IEEEoverridecommandlockouts                              
\def\FGPaperID{****} 

\title{\LARGE \bf
GaitPT: Skeletons Are All You Need For Gait Recognition}

\author{\parbox{16cm}{\centering
   {\large Andy Cătrună, Adrian Cosma, Emilian Rădoi}\\
   {\normalsize
   Faculty of Automatic Control and Computer Science, University Politehnica of Bucharest, Romania}
   {\tt\small catruna.andy@gmail.com, ioan\_adrian.cosma@upb.ro, emilian.radoi@upb.ro}
   \\}
}

\begin{document}

\ifFGfinal
\thispagestyle{empty}
\pagestyle{empty}
\else
\author{Anonymous FG2024 submission\\ Paper ID \FGPaperID \\}
\pagestyle{plain}
\fi

\maketitle

\begin{abstract}
The analysis of patterns of walking is an important area of research that has numerous applications in security, healthcare, sports and human-computer interaction. Lately, walking patterns have been regarded as a unique fingerprinting method for automatic person identification at a distance.  In this work, we propose a novel gait recognition architecture called Gait Pyramid Transformer (GaitPT) that leverages pose estimation skeletons to capture unique walking patterns, without relying on appearance information. GaitPT adopts a hierarchical transformer architecture that effectively extracts both spatial and temporal features of movement in an anatomically consistent manner, guided by the structure of the human skeleton. Our results show that GaitPT achieves state-of-the-art performance compared to other skeleton-based gait recognition works, in both controlled and in-the-wild scenarios. GaitPT obtains 82.6\% average accuracy on CASIA-B, surpassing other works by a margin of 6\%. Moreover, it obtains 52.16\% Rank-1 accuracy on GREW, outperforming both skeleton-based and appearance-based approaches.
\end{abstract}

\section{Introduction}
 Person re-identification is a long-standing problem in the field of computer vision and biometrics, that has prevalent real-world applications in areas such as public monitoring, security systems, and access control \cite{bryliuk2002access,jain2012biometric}. Currently, the most widespread methods of person identification typically involve processing biometric identifiers such as facial features \cite{parkhi2015deep}, fingerprints \cite{minaee2019fingernet}, or iris patterns \cite{arora2018computer}. While these methods are effective in controlled environments, they are heavily reliant on the active cooperation of the subjects, which is often not feasible in real-world scenarios, especially in public spaces. Facial recognition systems require the face of the individual to be fully visible under reasonably high resolution to work reliably \cite{li2018face}, and methods that rely on clothing matching \cite{ye2021deep} are bound to be ineffective across multiple days.

Gait analysis, the study of a person's unique walking pattern, has emerged as a promising alternative for person recognition \cite{singh2018vision}. Its study has gained significant attention \cite{gait-survey} as walking and movement have the potential to be used as a unique fingerprint enabling individual recognition in a wide range of uncooperative scenarios. Gait has been used so far to identify individuals \cite{sepas2022deep}, demographics \cite{lu2010gait,catruna2021face} 
 and estimate emotions \cite{emotion_estimation:Randhavane}.

Most approaches for gait-based person recognition have relied on the use of human silhouettes \cite{gait_set:Chao, gait_part:Fan,lin2022gaitgl}, which contain both gait and appearance information. The use of silhouettes raises major privacy concerns \cite{prabhakar2003biometric}, as they encode appearance information such as body composition, clothing and hairstyle. It is unclear how much the appearance information contained in silhouettes contributes to the successful recognition of a person on top of the gait information \cite{zhang2022realgait}, which by definition implies temporal variation in the form of movement. For instance, Xu et al. \cite{xu2021real} were able to construct a gait model that estimates the gender of an individual using only a single silhouette image. Such a model cannot be considered a gait-processing model, since it does not encode any form of temporal variation.

The use of human skeletons extracted with pretrained pose estimation models \cite{alphapose,xu2022vitpose} was proposed as a way to obtain gait information containing predominantly movement data \cite{li2020jointsgait,teepe2021gaitgraph,cosma2022learning} and removing all appearance information except for limb length. This approach makes gait recognition more privacy-friendly in comparison to sillhouete-based analysis. Methods typically use graph networks \cite{yan2018spatial,liu2020disentangling} to process human skeleton sequences, borrowing methods from human action recognition tasks. Concurrently, there has been recent growing interest in the use of vision transformers for computer vision tasks \cite{dosovitskiy2020image, wang2021pyramid}. Models based on the transformer architecture \cite{vaswani2017attention} have achieved state-of-the-art results on a wide range of problems, including image classification \cite{liu2021swin}, object detection \cite{carion2020end}, and segmentation \cite{xie2021segformer}. However, their application to gait-based person recognition remains relatively underdeveloped, with limited success reported in literature \cite{cosma2022learning,cosma23gaitvit}. 

Inspired by the recent advances in vision transformers \cite{wang2021pyramid,liu2021swin}, we propose a hierarchical transformer model for skeleton gait recognition, which we call GaitPT. It is designed to capture both spatial and temporal information of human movement in an anatomically consistent manner, by modelling the micro and macro movements of individual joints, iteratively combining them until the motion of the entire body is processed. The hierarchical structure of our model enables a gradual abstraction of the gait information, resulting in a more effective representation of the full movement pattern compared to previous graph-based \cite{li2020jointsgait,teepe2022towards} and transformer-based \cite{cosma2022learning} approaches.

We evaluate our model on three popular gait recognition benchmarks: CASIA-B benchmark \cite{CASIA:Yu}, a restrictive dataset captured in controlled scenarios, GREW \cite{GREW:zhu} one of the largest gait recognition benchmark in the wild, and Gait3D \cite{zheng2022gait3d} an in-the-wild dataset containing uncooperative and erratic gait sequences. Our experiments show that GaitPT outperforms previous skeleton-based state-of-the-art methods \cite{teepe2021gaitgraph,cosma2022learning,lima2021simple,li2020jointsgait,liao2020model}. The results indicate that our proposed hierarchical transformer architecture is a promising approach for gait-based person recognition.

This paper makes the following contributions: 
\begin{itemize}
    \item We propose Gait Pyramid Transformer (GaitPT), a novel skeleton-based gait processing architecture that achieves state-of-the-art results for skeleton gait recognition. We use anatomical priors in designing spatial and temporal attention blocks, which enable the hierarchical processing of human movement.
    
    \item We perform an extensive evaluation for gait recognition performance in both laboratory-controlled scenarios, as well as in realistic surveillance scenarios of erratic and uncooperative walking. 
    We evaluate our architecture on CASIA-B \cite{CASIA:Yu}, on GREW \cite{GREW:zhu} and on Gait3D \cite{zheng2022gait3d}, three popular gait recognition benchmarks, obtaining an average accuracy increase of 6\% over previous state-of-the-art methods.

    \item We conduct an ablation study on the architectural choices of GaitPT with the most impact on performance. In addition, we show that, for skeleton-based gait recognition, downstream accuracy is highly correlated with the upstream performance of the pose estimation model - the use of a state-of-the-art pose estimation model can result in upwards of 20\% accuracy gain.
    
\end{itemize}

\section{Related Work}
The approaches for gait-based person recognition are classified into two main categories: appearance-based \cite{person_identification:Wu,gait_set:Chao,gait_part:Fan,lin2022gaitgl} and model-based \cite{liao2020model,pose_based_recognition:Lima,teepe2021gaitgraph,teepe2022towards,cosma2021wildgait,cosma2022learning}. Recent appearance-based solutions process sequences of human silhouettes, utilize background subtraction \cite{zivkovic2004improved} or instance segmentation \cite{mask_rcnn:He,chen2019hybrid} models. Model-based approaches attempt to fit a custom model, such as an anatomical human model, over detected individuals across video frames. Generally, the anatomical model is the human skeleton obtained from pretrained human pose estimation networks \cite{alphapose,sun2019deep,OpenPose:Cao,xu2022vitpose}. 

\subsection{Appearance-based Approaches}
Efforts in appearance-based methods have utilized sequences of silhouettes that are processed by a convolutional network \cite{7533144,gait_set:Chao,gait_part:Fan,gln,lin2022gaitgl}. For instance, the authors of GaitSet \cite{gait_set:Chao} propose an approach that represents gait as an unordered set of silhouettes. By adopting this representation, they argue that their approach is more adaptable to varying frame arrangements and different walking directions and variations compared to vanilla silhouette sequences. GaitSet leverages convolution layers to extract image-level features from each silhouette and uses Set Pooling to aggregate them into a set-level feature. To produce the final output, the authors employ a custom version of Horizontal Pyramid Matching \cite{fu2019horizontal}. Fan et al. \cite{gait_part:Fan} observed that different parts of the human silhouette exhibit unique spatio-temporal patterns, and thus require their specific expression. The authors proposed an architecture with a specialized type of convolution with a restricted receptive field which enables processing specific parts of the body, more specifically the head, the upper body, and the legs. GLN \cite{gln} utilizes a convolutional feature pyramid to learn compact gait representations. According to Lin et al. \cite{lin2022gaitgl}, existing methods for gait analysis face a trade-off between capturing global versus local information. Methods that focus on global features may overlook important local details such as small movements, while methods that extract local features may not fully capture the relationships between them, losing the global context in the process. To address this limitation, the authors propose the GaitGL architecture, which incorporates a two-stream network that captures both global and local gait features simultaneously. 

Both GaitSet \cite{gait_part:Fan}, GLN \cite{gln} and GaitGL \cite{lin2022gaitgl} recognize that efficiently processing the hierarchical relationship between local and global features is an essential characteristic of a performant gait recognition method. Our GaitPT architecture builds upon this idea, utilizing an anatomically informed model based on the human skeleton to process the micro and macro movements of a walking individual. We chose to use a model-based approach since skeletons encode mainly movement data and can be considered privacy-friendly while the use of silhouettes more invasive as the recognition performance can be influenced by appearance features.

\subsection{Model-based Approaches}
Processing a sequence of human skeletons generally entails the use of a graph network, developed primarily for skeleton action recognition problems. Networks such as ST-GCN \cite{yan2018spatial} and MS-G3D \cite{liu2020disentangling} have been repurposed for gait recognition in several works \cite{li2020jointsgait,teepe2021gaitgraph,teepe2022towards,cosma2021wildgait}. For instance, Li et al. \cite{li2020jointsgait} propose JointsGait, an architecture that leverages graph convolutions. They utilize the ST-GCN architecture \cite{yan2018spatial} to capture the spatio-temporal features from sequences of skeletons. Furthermore, they use a Joints Relationship Pyramid Mapping which maps the extracted features into a more discriminative space by exploiting the areas of the body which naturally work together. Teepe et al. \cite{teepe2021gaitgraph,teepe2022towards} employ the ResGCN \cite{song2020stronger} architecture to capture spatio-temporal features from skeleton sequences. They train their architecture with the supervised contrastive objective \cite{khosla2020supervised} and leverage multiple augmentation techniques such as flipping, mirroring, and adding noise to the data. Cosma and Radoi \cite{cosma2021wildgait} leverage surveillance footage to create a large-scale skeleton dataset, which they use to pretrain an ST-GCN architecture in a self-supervised manner, with good downstream transfer capabilities. Some works defer to using plain CNN / MLP models to process the sequence of skeletons. PoseGait, introduced by Liao et al. \cite{liao2020model} involves extracting human keypoints from each video frame and computing hand-crafted features, including the angle of the limbs, the length of the limbs and joint motion, to facilitate the extraction of gait features. The model utilizes a CNN to model the temporal relationship between these features across frames, enabling effective recognition of gait patterns. Lima et al. \cite{pose_based_recognition:Lima} utilize a multilayer perceptron on individual skeletons to capture the spatial information. They use skeleton normalization based on the corresponding neck coordinate to remove information about the position in the image. The final embedding is obtained by temporally aggregating all the spatial features extracted with the MLP. 

With the prevalent use of the transformer model \cite{vaswani2017attention} across most areas of deep learning, some works are exploring the use of transformer models for gait recognition \cite{cosma2022learning,cosma23gaitvit}. GaitFormer \cite{cosma2022learning} was the first application of the transformer architecture to gait analysis problems to process sequences of silhouettes. However, the authors only employ temporal attention by flattening each skeleton, ignoring spatial relationships and low-level movements.

Different from Cosma et al. \cite{cosma2022learning}, we utilize an anatomically informed model to construct spatial and temporal attention blocks, such that micro and macro movements are processed hierarchically. Our experiments show that our architecture is effective in modelling gait, outperforming previous state-of-the-art graph-based methods as well as transformer-based methods by a large margin.

\section{Method}
\begin{figure*}
\begin{center}
    \includegraphics[width=\textwidth]{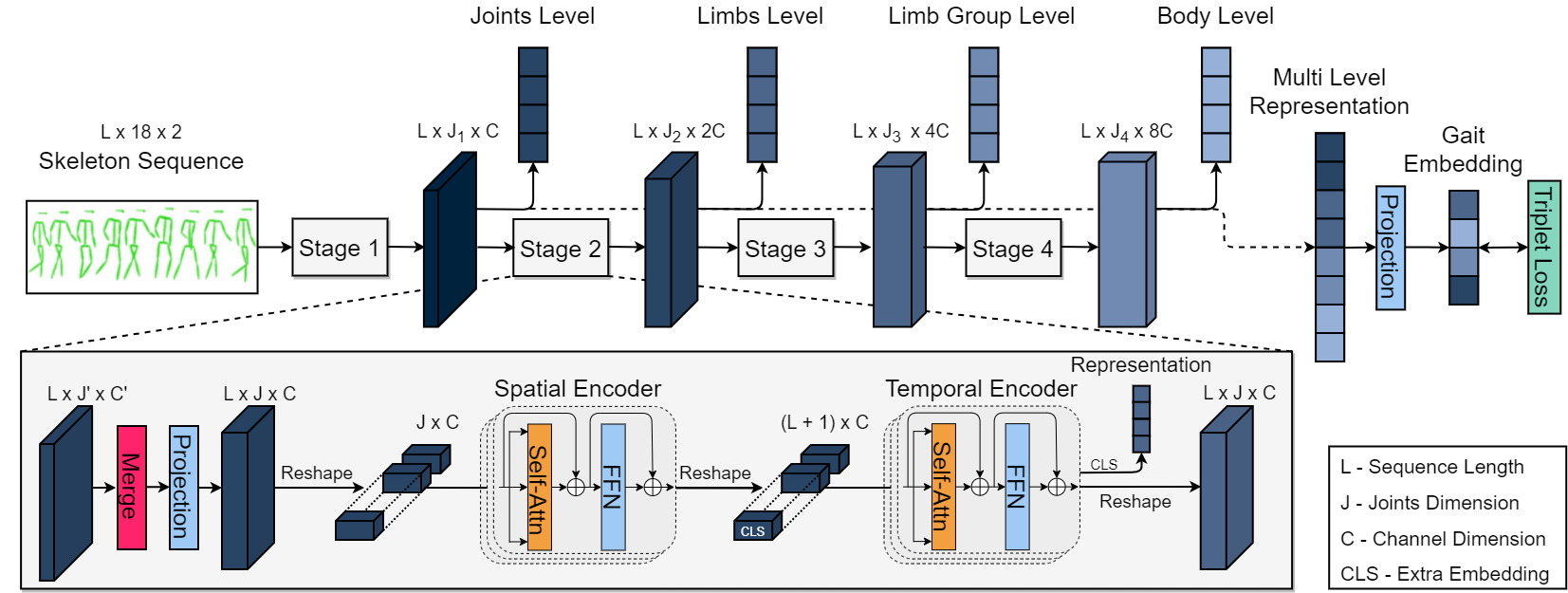}    
\end{center}
  \caption{Overview of the GaitPT architecture. The model uses spatial and temporal attention to incrementally learn the natural motion of the human body. In Stage 1 it computes the spatio-temporal interactions at the joint level, in Stage 2 at the limb level, in Stage 3 at the group of limbs level and in the final Stage computes the interaction between full skeletons. The representations obtained from every stage are combined into a gait embedding that captures discriminative features at all levels of movement.}
  \label{fig:pgt}
\end{figure*}

\subsection{Preliminaries}
In the initial transformer architecture \cite{vaswani2017attention}, the encoder receives as input a sequence of elements ($X \in R^{n \times d}$ where $n$ - number of items, $d$ - embedding dimension) that are multiplied with 3 learnable matrices to compute the Queries ($Q \in R^{n \times d_q}$), the Keys ($K \in R^{n \times d_k}, d_k = d_q$), and Values ($V \in R^{n \times d_v}$). The self-attention operation captures the interactions between the items in the sequence. It can be written as:
\begin{equation}
    Attention(Q,K,V) = Softmax(QK^{T} / \sqrt{d_k}) V
    \label{eq:attention}
\end{equation}
Vaswani et al. \cite{vaswani2017attention} note that computing multiple interactions between the elements of a sequence improves performance. Consequently, Multi-head Self-Attention (MSA) is employed. It is computed as:
\begin{equation}
    MSA(Q,K,V) = Concat(head_1,...,head_h) W^O
\end{equation}
where $head_i = Attention(QW^Q_i, KW^K_i, VW^V_i)$.

The transformer encoder layer computes the following operations to obtain the output $z_{l+1}$ from the sequence of tokens received as input ($z_l \in R^{n \times d}$):
\begin{equation}
    \begin{aligned}
    z'_l = LN(z_l + MSA(z_l))\\
    z_{l+1} = LN(z'_l + FFN(z'_l))
    \end{aligned}
    \label{eq:encoder_layer}
\end{equation}
where LN stands for LayerNorm and FFN for FeedForward Network. Residual connections are also employed after the MSA block and the FFN block.

Vision Transformers employ variations of the transformer encoder which use the LayerNorm module before the Self-Attention and FeedForward layer. This computation can be described as:
\begin{equation}
    \begin{aligned}
    z'_l = z_l + MSA(LN(z_l))\\
    z_{l+1} = z'_l + FFN(LN(z'_l))
    \end{aligned}
    \label{eq:vit_layer}    
\end{equation}


In line with other works \cite{cosma2022learning,teepe2021gaitgraph} in skeleton-based gait analysis, we operate on sequences of human skeletons extracted from RGB images using pretrained pose estimation models. Human pose estimators return a pose $\boldsymbol{p}$ which represents a set of 17 coordinates of the most important joints of the body. These can be defined as $\boldsymbol{p} = \{j^1,j^2...,j^{17}\}$, where $j^i$ stands for the $i$th joint in the pose and contains the coordinates $(x^i, y^i)$. We insert an additional 18th coordinate by duplicating the nose coordinate for symmetry purposes in our architecture. A walking sequence is obtained by concatenating consecutive poses $Z=\{\boldsymbol{p_1},\boldsymbol{p_2},...,\boldsymbol{p_n}\}, Z \in R^{n \times 18 \times 2}$.

\subsection{Gait Pyramid Transformer}
A high level overview of the proposed architecture GaitPT is shown in Figure \ref{fig:pgt}. Our model operates in a hierarchical manner by first computing the movement of individual joints, followed by individual limbs, then groups of limbs, and finally the full body. This approach introduces an inductive bias to the architecture, enabling the model to capture anatomically informed unique walking patterns. The GaitPT architecture builds upon the spatial and temporal attention-based modules from the work of Plizzari et al. \cite{plizzari2021spatial}. However, while their architecture only computes interactions at the joint-level, our model uses a hierarchical approach similar to that of PVT \cite{wang2021pyramid} to capture both local and global unique gait features. The main building blocks of GaitPT are the Spatial Encoder, the Temporal Encoder, and the Joint Merging module.

\begin{figure}
\begin{center}
    \includegraphics[width=\linewidth]{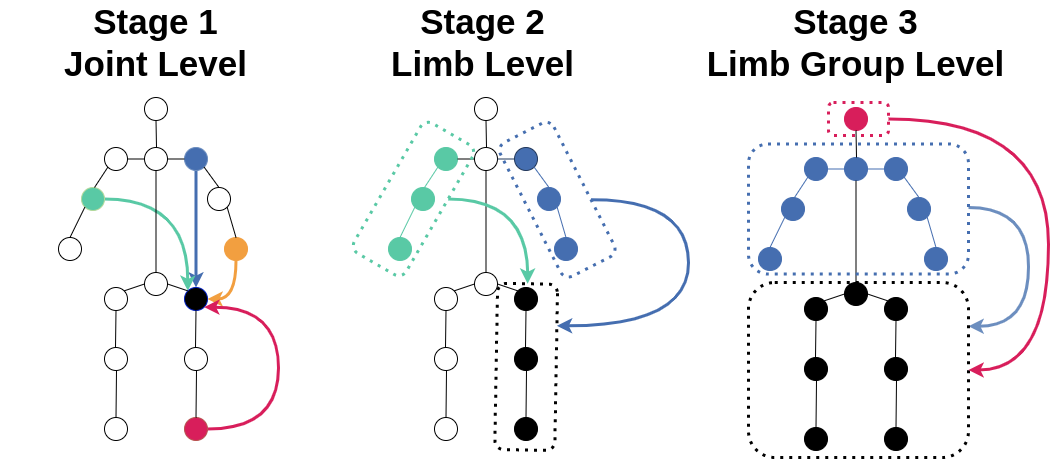}    
\end{center}
  \caption{Visualization of the Spatial Attention across all stages in the GaitPT architecture. Spatial Attention is applied across multiple joints / limbs of the body in the same time step. Spatial Attention is performed at the joint level at Stage 1, at the limb level at Stage 2, and at the level of groups of limbs at Stage 3.}
  \label{fig:spatial}
\end{figure}

\textbf{Spatial Attention} computes the interactions between the joints of individual skeletons by considering each pose as a separate sequence. Having the input sequence $Z=\{\boldsymbol{p_1},\boldsymbol{p_2},...,\boldsymbol{p_n}\}$, spatial attention uses the same computations as in Equation \ref{eq:vit_layer} for each $\boldsymbol{p_i}, i=1..n$. This is done with a \textit{Reshape} operation that transforms $Z$ into a list of individual poses. Figure \ref{fig:spatial} shows a visualization of the interactions computed by the spatial encoder of the GaitPT architecture across all stages. In Stage 4 there is no spatial attention done as the partitioning at that level is at the full body level. 

\begin{figure}
\begin{center}
    \includegraphics[width=0.85\linewidth]{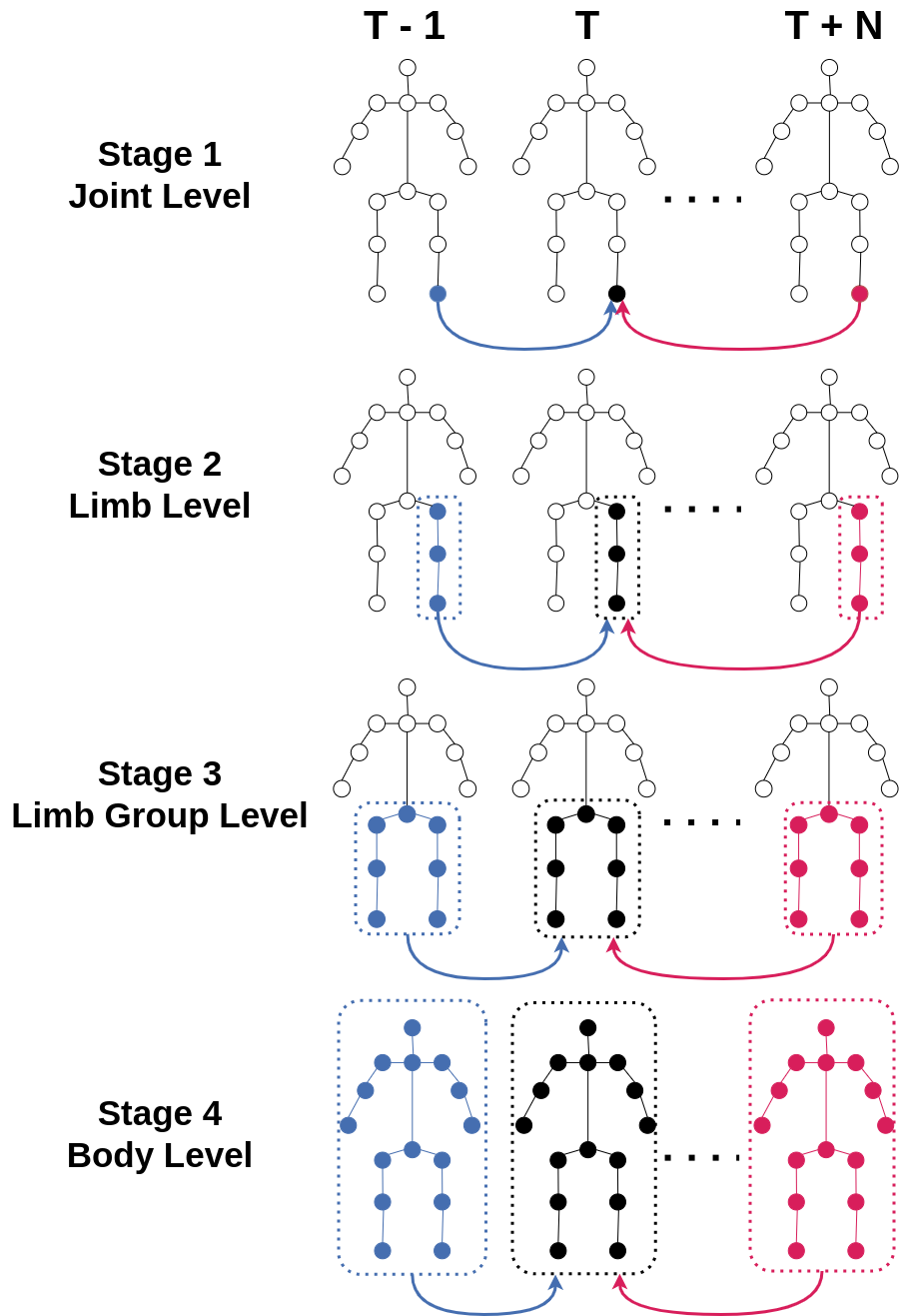}   
\end{center}
  \caption{Visualization of the Temporal Attention across all stages in the GaitPT architecture. Temporal Attention is applied to the same joints / limbs across different time-steps. Initially, Temporal Attention is performed at the joint level at Stage 1, at the limb level at Stage 2, at the level of groups of limbs at Stage 3 and at the whole body level in Stage 4.}
  \label{fig:temporal}
\end{figure}

\textbf{Temporal Attention} computes the relation between the same feature vector at different time steps. Given a walking sequence $Z=\{\boldsymbol{p_1},\boldsymbol{p_2},...,\boldsymbol{p_n}\}$, where $\boldsymbol{p_i}= \{j^1_i,j^2_i...,j^m_i\}$ (m - number of feature vectors in pose which differs based on the stage) temporal attention utilizes the same computations as in Equation \ref{eq:vit_layer} with sequences of the form $z=\{j^i_1, j^i_2,...,j^i_n\}$, where $i=1..m$. In other words, it receives as input sequences of the same feature vector at all the time steps from the gait sequence. This is also made possible with a \textit{Reshape} operation that transforms the output of the Spatial Attention module into lists that contain the evolution of individual embeddings (corresponding to joints or groups of joints) in time. Figure \ref{fig:temporal} shows a visualization of the interactions computed across time steps between the same feature vectors.

The \textbf{Joint Merging} module takes as input a pose or intermediary feature map and combines groups of vectors based on their anatomical relationship. These groups are specifically chosen so that they correspond to joints that naturally work together in the human motion. As shown in Figure \ref{fig:temporal}, the merging module initially combines feature vectors corresponding to individual joints to create feature vectors corresponding to limbs. For instance, the embedding of the left shoulder is merged with the embeddings of the left elbow and of the left wrist to form the feature vector corresponding to the whole left arm. Subsequently, feature vectors associated with individual limbs are united, such as the merging of the left and right legs to form the lower body region. Lastly, all remaining feature vectors are merged into a single vector that encapsulates the information of the entire body.

In total the GaitPT architecture has 4 stages. In the first stage, spatio-temporal interactions are computed at the joint level. In the second stage, after the Merging module, the spatial and temporal encoders receive as input feature vectors that correspond to limbs. The third stage introduces the limb group level of interaction. In this stage the limbs are combined into the following groups: head area, upper body area, and lower body area. This partitioning scheme is similar to that of GaitPart \cite{gait_part:Fan}. However, while their approach splits the image based on manually designed values which may not always yield these 3 parts specifically, our partitioning precisely divides the body into these 3 groups as it is based on the pose estimation skeleton. We study multiple grouping schemes in Section 4.4 to obtain the most suitable candidate. In the final stage, only temporal attention is performed at the level of the full body.  

Given a sequence of vectors ($Z_l \in R^{n \times d}$) as input to a GaitPT layer, the output $Z_{l+1}$ is computed as:
\begin{equation}
    \begin{aligned}
    Z'_{l} = Reshape(JointMerge(Z_{l})) \\
    Z''_{l} = Reshape(SpatialAttention(Z'_{l})) \\
    Z_{l+1} = Reshape(TemporalAttention(Z''_{l})) \\
    \end{aligned}
\end{equation}
The Spatial and Temporal Attention operations are exactly the same as the ones in Equation \ref{eq:vit_layer}. The Reshape operation is utilized to obtain the correct token sequence for the corresponding encoder or for the output of the layer. In line with ViT architectures \cite{dosovitskiy2020image, touvron2021going, wang2021pyramid} we incorporate an extra class token for each temporal encoder to obtain an embedding that captures discriminative features. The class outputs obtained from each stage are aggreagted and projected with a linear layer to a lower dimension to obtain the final embedding.

\subsection{Implementation Details}
To obtain gait embeddings that capture discriminative features of walking sequences, we employ the triplet loss objective \cite{schroff2015facenet} with a margin of 0.02 to train the GaitPT architecture. The training objective takes as input an embedding triplet  consisting of an anchor embedding $(a)$, a positive embedding $(p)$, and negative embedding $(n)$, yielding a triplet $(a,p,n)$. The objective maximizes the distance between the anchor and negative embeddings, while minimizing the distance between the anchor and positive embedding, constructing an euclidean space where inference can be done based on nearest neighbors. The triplet loss is defined as:
\begin{equation}
L(a,p,n) = d(a,p) - d(a,n) + m
\end{equation}
where $m$ is the margin which enforces a minimum distance between positive and negative embeddings and $d$ is the distance function.

Each stage linearly projects the input feature vectors to a higher dimension. After Stage 1 the embedding dimension $C$ is equal to 32 and is multiplied by a factor of 2 in every following stage. The spatial and temporal encoders at each stage consist of 3 transformer encoder blocks, each with internal feed-forward dimension of $4 * C$. Each stage outputs both the feature maps for the next stage and a representation of the current level of movement. This representation is an embedding computed in the extra CLS token of the temporal encoder. The final output of the architecture is an embedding of size 256 which is obtained by combining the representations from all 4 stages through concatenation and linear projection. This approach ensures that the embedding contains discriminative features at all levels of movement. 

The training is performed using the AdamW optimizer \cite{loshchilov2017decoupled} and a cyclic learning rate schedule \cite{smith2017cyclical}, which starts from a minimum of 0.0001 and reaches a maximum of 0.01, with an exponential decay rate of 0.995 and a step size of 15. 
We used a single NVIDIA A100 with 40GB of VRAM for our experiments. A complete training run takes 1.5h for CASIA-B, 12h for GREW and 3h for Gait3D. In total, GaitPT has 4M parameters.

\section{Experiments and Results}
To evaluate the performance of our model, we chose two different scenarios: a controlled laboratory environment and a more realistic, "in-the-wild" scenario. CASIA-B benchmark \cite{CASIA:Yu} is used to evaluate our model against other state-of-the-art approaches in gait recognition for controlled scenarios, while GREW \cite{GREW:zhu} and Gait3D \cite{zheng2022gait3d} are used to test whether our architecture generalizes to unconstrained environments. CASIA-B, GREW, and Gait3D are available upon request, and can be accessed through a release agreement.

\begin{table*}[hbt!]
    \caption{GaitPT comparison to other skeleton-based architectures on CASIA-B. We report the average recognition accuracy for individual probe angles excluding identical-view cases. For our model, we show the mean and standard deviation computed across 3 runs. GaitPT obtains an average improvement of over 6\% mean accuracy compared to the previous state-of-the-art.}
    \label{tab:casia_sota}
    \begin{center}
    \resizebox{\textwidth}{!}{
    \begin{tabular}{ c | l | ccccccccccc | c }
         \textbf{Scenario} & \textbf{Method} & 0$^{\circ}$  &    18$^{\circ}$  &    36$^{\circ}$  &    54$^{\circ}$  &    72$^{\circ}$  &    90$^{\circ}$  &    108$^{\circ}$ &    126$^{\circ}$ &   144$^{\circ}$ &    162$^{\circ}$ &    180$^{\circ}$ & \textbf{Mean} \\
        \midrule
        \multirow{ 7}{*}{NM\#5-6} & PoseGait \cite{liao2020model} & 48.5 & 62.7 & 66.6 & 66.2 & 61.9 & 59.8 & 63.6 & 65.7 & 66 & 58 & 46.5 & 60.5 \\
        & JointsGait \cite{li2020jointsgait} & 68.1 & 73.6 & 77.9 & 76.4 & 77.5 & 79.1 & 78.4 & 76 & 69.5 & 71.9 & 70.1 & 74.4 \\
 
        & PoseFrame \cite{lima2021simple} & 66.2 & \textbf{93.9} & 88.5 & 56.1 & 79.7 & \textbf{98.0} & \textbf{98.6} & \textbf{99.3} & 81.8 & 80.4 & 70.3 & 83.0 \\
        & GaitFormer \cite{cosma2022learning} & 71.0 & 74.7 & 77.5 & 77.1 & 77.4 & 77.3 & 78.1 & 71.5 & 79.4 & 74.0 & 66.5 & 74.9 \\
        & GaitGraph \cite{teepe2021gaitgraph} & 85.3 & 88.5 & 91.0 & 92.5 & 87.2 & 86.5 & 88.4 & 89.2 & 87.9 & 85.9 & 81.9 & 87.7  \\
        & GaitGraph2 \cite{teepe2022towards} & 78.5 & 82.9 & 85.8 & 85.6 & 83.1 & 81.5 & 84.3 & 83.2 & 84.2 & 81.6 & 71.8 & 82.0   \\
        & \textbf{GaitPT (Ours)} & \textbf{93.5} \small{$\pm$ 0.5} & 92.0 \small{$\pm$ 1.0} & \textbf{94.3} \small{$\pm$ 0.6} & \textbf{93.9} \small{$\pm$ 1.7} & \textbf{92.5} \small{$\pm$ 0.7} & 92.3 \small{$\pm$ 1.3} & 92.4 \small{$\pm$ 0.9} & 92.9 \small{$\pm$ 1.2} & \textbf{92.9} \small{$\pm$ 1.4} & \textbf{92.5} \small{$\pm$ 1.6} & \textbf{86.5} \small{$\pm$ 1.5} & \textbf{92.3} \small{$\pm$ 0.7}  \\
        \midrule
        
        \multirow{ 7}{*}{BG\#1-2} & PoseGait \cite{liao2020model} & 29.1 & 39.8 & 46.5 & 46.8 & 42.7 & 42.2 & 42.7 & 42.2 & 42.3 & 35.2 & 26.7 & 39.6 \\
        & JointsGait \cite{li2020jointsgait} & 54.3 & 59.1 & 60.6 & 59.7 & 63 & 65.7 & 62.4 & 59 & 58.1 & 58.6 & 50.1 & 59.1  \\
        & PoseFrame \cite{lima2021simple} & 48.6 & 69.6 & 56.1 & 41.9 & 56.8 & \textbf{84.5} & \textbf{80.4} & \textbf{83.1} & 65.5 & 58.1 & 48.0 & 63.0 \\
        & GaitFormer \cite{cosma2022learning} & 63.2 & 64.4 & 66.0 & 60.7 & 62.3 & 62.0 & 61.2 & 54.2 & 55.9 & 61.3 & 51.3 & 60.2 \\
        & GaitGraph \cite{teepe2021gaitgraph} & 75.8 & 76.7 & 75.9 & 76.1 & 71.4 & 73.9 & 78.0 & 74.7 & 75.4 & 75.4 & 69.2 & 74.8  \\
        & GaitGraph2 \cite{teepe2022towards} & 69.9 & 75.9 & 78.1 & 79.3 & 71.4 & 71.7 & 74.3 & 76.2 & 73.2 & 73.4 & 61.7 & 73.2  \\
        & \textbf{GaitPT (Ours)} & \textbf{83.4} \small{$\pm$ 4.0} & \textbf{80.5} \small{$\pm$ 3.9} & \textbf{83.4} \small{$\pm$ 1.9} & \textbf{82.1} \small{$\pm$ 1.4} & \textbf{76.2} \small{$\pm$ 2.0} & 75.7 \small{$\pm$ 2.0} & 78.3 \small{$\pm$ 3.0} & 79.1 \small{$\pm$ 3.4} & \textbf{80.8} \small{$\pm$ 2.6} & \textbf{82.0} \small{$\pm$ 1.5} & \textbf{74.9} \small{$\pm$ 3.0} & \textbf{79.7} \small{$\pm$ 2.5}   \\
        \midrule
        
        \multirow{ 7}{*}{CL\#1-2} & PoseGait \cite{liao2020model} & 21.3 & 28.2 & 34.7 & 33.8 & 33.8 & 34.9 & 31 & 31 & 32.7 & 26.3 & 19.7 & 29.8 \\
        & JointsGait \cite{li2020jointsgait} & 48.1 & 46.9 & 49.6 & 50.5 & 51 & 52.3 & 49 & 46 & 48.7 & 53.6 & 52 & 49.8  \\
        & PoseFrame \cite{lima2021simple} & 16.2 & 28.4 & 21.6 & 23.6 & 28.4 & 50.7 & 54.1 & 49.3 & 28.4 & 25.0 & 20.3 & 31.4 \\
        & GaitFormer \cite{cosma2022learning} & 47.1 & 45.6 & 44.6 & 45.3 & 48.8 & 46.2 & 52.3 & 41.5 & 43.8 & 46.0 & 45.7 & 46.1 \\
        & GaitGraph \cite{teepe2021gaitgraph} & 69.6 & 66.1 & 68.8 & 67.2 & 64.5 & 62.0 & 69.5 & 65.6 & 65.7 & 66.1 & 64.3 & 66.3  \\
        & GaitGraph2 \cite{teepe2022towards} & 57.1 & 61.1 & 68.9 & 66 & 67.8 & 65.4 & 68.1 & 67.2 & 63.7 & 63.6 & 50.4 & 63.6   \\
        & \textbf{GaitPT (Ours)} & \textbf{76.0} \small{$\pm$ 2.0} & \textbf{77.5} \small{$\pm$ 1.5} & \textbf{76.5} \small{$\pm$ 3.4} & \textbf{77.6} \small{$\pm$ 1.6} & \textbf{73.3} \small{$\pm$ 2.8} & \textbf{76.6} \small{$\pm$ 1.4} & \textbf{77.0} \small{$\pm$ 3.0} & \textbf{75.3} \small{$\pm$ 2.0} & \textbf{73.4} \small{$\pm$ 3.0} & \textbf{76.3} \small{$\pm$ 3.4} & \textbf{74.8} \small{$\pm$ 1.8} & \textbf{75.8} \small{$\pm$ 2.0} \\
    \end{tabular}
    }
    \end{center}
\end{table*}

\subsection{Evaluation in Controlled Scenarios}

CASIA-B consists of walking video data from 124 unique subjects. Each subject has 6 normal (NM) walking sessions, 2 sessions while carrying a bag (BG), and 2 sessions while wearing a coat (CL). Each session consists of 11 videos recorded with multiple synchronized cameras to obtain various walking angles. In total, each subject has 110 videos. We follow the same partitioning of train and test as Teepe et al. \cite{teepe2021gaitgraph} where the first 74 subjects are utilized for training while the remaining 50 for testing. The evaluation protocol requires that the first 4 normal (NM\#1-4) sessions are utilized as the gallery samples. The remaining normal sessions (NM\#5-6) and all the bag carrying (BG\#1-2) and coat wearing (CL\#1-2) scenarios form 3 separate probe sets. Gallery-probe pairs are constructed based on each individual angle, excluding the same view scenario. For each gallery-probe pair, the accuracy of the model is measured individually, and the reported result for each specific probe angle is the average accuracy across all the gallery angles. We do minimal preprocessing consisting of normalizing the pose estimation coordinates by the width of the video frames. We also remove sequences which have less than 60 total frames which is in line with the work of Teepe et al. \cite{teepe2021gaitgraph, teepe2022towards}.

We present a comparative analysis of our model with other state-of-the-art skeleton-based approaches \cite{li2020jointsgait,lima2021simple,cosma2022learning,teepe2021gaitgraph,teepe2022towards}, that use qualitatively different types of architectures to model skeleton sequences. \textbf{PoseGait} \cite{lima2021simple} utilizes a simple MLP at the skeleton level and aggregates information across time. \textbf{JointsGait} \cite{li2020jointsgait} and \textbf{GaitGraph} \cite{teepe2021gaitgraph} are two graph-based models that utilize as a backbone network a ST-GCN \cite{yan2018spatial} and a ResGCN \cite{song2020stronger} respectively. \textbf{GaitGraph2} \cite{teepe2022towards} is a subsequent improvement to GaitGraph, which incorporates pre-computed features from the skeleton information such as the motion velocity, the bone length and the bone angle. \textbf{GaitFormer} \cite{cosma2022learning} is the only other transformer-based architecture for gait recognition, but only utilizes temporal attention. The authors evaluate their model on a different CASIA-B split than our benchmark. We utilize the open-source implementation and train the model under the same conditions as GaitPT.

In Table \ref{tab:casia_sota}, we showcase results on the CASIA-B dataset, with methods following the same evaluation protocol. Our proposed architecture, GaitPT, surpasses the previous state-of-the-art, GaitGraph \cite{teepe2021gaitgraph}, which utilizes graph convolutions for spatio-temporal feature extraction. GaitFormer \cite{cosma2022learning} lags behind other graph-based methods, proving that spatial attention is a crucial component in gait recognition. Our results demonstrate the effectiveness of a hierarchical approach to motion understanding in the context of gait recognition in controlled scenarios.

\subsection{Evaluation In the Wild}

As the GaitPT architecture achieves adequate recognition performance in controlled settings, we study its capabilities in more difficult real-world scenarios.

%

\begin{table}[hbt!]
    \caption{Comparison between GaitPT and other methods on the GREW benchmark in terms of Rank-1, Rank-5, Rank-10, and Rank-20 Recognition Accuracy. For our model, we show the mean and standard deviation computed across 3 runs. GaitPT manages to outperform by a significant margin both skeleton-based and appearance-based state-of-the-art methods for in-the-wild scenarios. Table adapted from \cite{GREW:zhu}.}
    \label{tab:grew_sota}
    \begin{center}
    \resizebox{\linewidth}{!}{
        \begin{tabular}{ l | c | c | c | c }
             \textbf{Method} & \textbf{R-1 Acc. (\%)}  & \textbf{R-5 Acc. (\%)} & \textbf{R-10 Acc. (\%)} & \textbf{R-20 Acc. (\%)} \\
            \midrule
            GEINet \cite{shiraga2016geinet} & 6.82 & 13.42 & 16.97 & 21.01 \\
            TS-CNN \cite{wu2016comprehensive} & 13.55 & 24.55 & 30.15 & 37.01 \\
            GaitSet \cite{gait_set:Chao} & 46.28 & 63.58 & 70.26 & 76.82  \\
            GaitPart \cite{gait_part:Fan} & 44.01 & 60.68 & 67.25 & 73.47
            \\
            PoseGait \cite{liao2020model} & 0.23 & 1.05 & 2.23 & 4.28
            \\
            GaitGraph \cite{teepe2021gaitgraph} & 1.31 & 3.46 & 5.08 & 7.51
            \\
            \textbf{GaitPT (Ours)} & \textbf{52.16} \small{$\pm$ 0.5} & \textbf{68.44} \small{$\pm$ 0.6} & \textbf{74.07} \small{$\pm$ 0.5} & \textbf{78.33} \small{$\pm$ 0.4}   \\
        \end{tabular}
    }
    \end{center}
\end{table}

\textbf{GREW} \cite{GREW:zhu} is one of the largest benchmarks for gait recognition in the wild, containing 26,000 unique identities and over 128,000 gait sequences. The videos for this dataset were obtained from 882 cameras in public spaces and the labelling was done by 20 annotators for over 3 months. GREW releases silhouettes, optical flow, 3D skeletons, and 2D skeletons for the recorded walking sequences.

We train the GaitPT architecture on the provided 2D skeletons which are normalized based on the dimensions of first image in the sequence. Based on the smallest relevant walking sequences of the dataset, we utilize a sequence length of 30 for both training and testing, which is in line with other works in gait recognition \cite{zheng2022gait3d}. As the GREW authors do not release the labels for the testing set, we report the results obtained on the public leaderboard\footnote{\url{https://codalab.lisn.upsaclay.fr/competitions/3409\#results}} for the GREW Competition.

Table \ref{tab:grew_sota} displays the comparison between GaitPT and other methods, including both skeleton-based and appearance-based approaches, on the GREW test set in terms of Rank-1, Rank-5, Rank-10, and Rank-20 accuracy. Rank-K accuracy computes the percentage of samples for which the correct label is among the top K predictions made by the model. The results of the other models are taken from the GREW \cite{GREW:zhu} paper. GaitPT outperforms skeleton-based approaches such as PoseGait \cite{liao2020model} and GaitGraph \cite{teepe2021gaitgraph} by over 50\% in Rank-1 Accuracy. Moreover, it manages to outperform appearance-based state-of-the-art methods such as GaitSet \cite{gait_set:Chao} and GaitPart \cite{gait_part:Fan} by approximately 6\% and 8\% in terms of Rank-1 Accuracy. These results demonstrate the capabilities of GaitPT to generalize in unconstrained settings and the fact that skeleton-based data can be advantageous for in-the-wild scenarios.

\textbf{Gait3D} \cite{zheng2022gait3d} is a large-scale dataset obtained in unconstrained settings consisting of 4000 unique subjects and over 25,000 walking sequences. The dataset includes 3D meshes, 3D skeletons, 2D skeletons, and silhouette images obtained from all recorded sequences.

We train our architecture using the provided 2D skeletons obtained through HRNet \cite{sun2019deep}, following the same evaluation protocol as Zheng et al. \cite{zheng2022gait3d} in which 3000 subjects are placed in the training set and the remaining 1000 in the gallery-probe sets. Similarly to the evaluation on CASIA-B, we normalize the 2D skeletons based on the dimensions of the image. In line with the methodology employed by the authors of Gait3D, we utilize a sequence length of 30 during training and testing. 

Table \ref{tab:gait3d_sota} displays the results of the Gait3D test set in terms of Rank-1 and Rank-5 accuracy. Results for PoseGait \cite{lima2021simple} and GaitGraph \cite{teepe2021gaitgraph} are directly taken from Gait3D \cite{zheng2022gait3d} results benchmark. We train both GaitPT and GaitFormer \cite{cosma2021wildgait} using this approach with the same hyperparameters detailed in Section 3.3. GaitPT obtains an average increase of 6.25\% in terms of rank-1 accuracy and an average improvement of 8.58\% in rank-5 accuracy compared to the previous skeleton-based state-of-the-art. These results demonstrate that our architecture generalizes effectively even in real-world scenarios where accurate recognition is challenging. 

\begin{table}[hbt!]
    \caption{Comparison between our architecture and other skeleton-based models on the Gait3D benchmark in terms of Rank-1 and Rank-5 Recognition Accuracy. All methods are trained on the same pose estimation data. For our model, we show the mean and standard deviation computed across 3 runs. GaitPT obtains an average increase of 6.25\% in Rank-1 Accuracy over previous state-of-the-art.}
    \label{tab:gait3d_sota}
    \begin{center}
    \resizebox{\linewidth}{!}{
        \begin{tabular}{ l | c | c }
             \textbf{Method} & \textbf{R-1 Accuracy (\%)}  & \textbf{R-5 Accuracy (\%)} \\
            \midrule
            PoseGait \cite{liao2020model} & 0.24 & 1.08 \\
            GaitGraph \cite{teepe2021gaitgraph} & 6.25 & 16.23 \\
            GaitFormer \cite{cosma2022learning} & 6.94 & 15.56  \\
            \textbf{GaitPT (Ours)} & \textbf{13.19} \small{$\pm$ 0.7} & \textbf{24.14} \small{$\pm$ 2.1}   \\
        \end{tabular}
    }
    \end{center}
\end{table}

\subsection{Ablation on GaitPT Stages }

To understand the capabilities of the GaitPT architecture and the necessity of each stage, we perform an ablation study in which we train the model with different stages activated. Table \ref{tab:ablation} displays the performance of the architecture for each stage configuration. The results show that each stage is crucial in modelling increasingly complex movement patterns to achieve good recognition performance. There is a significant (pairwise Welch's t-test $p < 0.05$) increase in performance after the addition of each stage. As Stage 4 does not have any type of spatial attention by itself, its performance gets improved tremendously with the addition of Stage 1 which models spatial interactions. Stages 2 and 3 add incremental performance by increasing the complexity of movement combinations. 

\begin{table}[hbt!]
    \caption{Ablation study for the GaitPT architecture on CASIA-B. We show mean accuracy for 10 runs, alongside standard deviation. For brevity, we only report the average accuracy across all walking angles and variations. There is a significant (pairwise Welch's t-test $p < 0.05$) increase in performance with each added stage.}
    \label{tab:ablation}
    \begin{center}
    \resizebox{\linewidth}{!}{
        \begin{tabular}{c | c | c | c | c}
             \textbf{Stage 1} & \textbf{Stage 2}  & \textbf{Stage 3} & \textbf{Stage 4} & \textbf{Mean Accuracy} \\
             \hline
             - & - &  - & \checkmark & 57.42 \small{$\pm$ 1.5} \\
             \checkmark & - & - & \checkmark & 73.48 \small{$\pm$ 3.8} \\
             \checkmark & \checkmark & - & \checkmark & 76.80 \small{$\pm$ 2.5} \\
             \checkmark & \checkmark & \checkmark &  \checkmark & \textbf{78.85} \small{$\pm$ 1.4} \\
        \end{tabular}
    }
    \end{center}
\end{table}

Among all stages in the GaitPT architecture, only the third stage lacks a clear and anatomically-informed approach for merging limb embeddings into groups that naturally work together during movement. Unlike Stage 1, which functions at the joint level, and Stage 2, which operates at the limb level, the limb group tokens of Stage 3 cannot be easily defined based on human anatomy. Consequently, we conducted experiments with various partitioning schemes that use different strategies for merging limb feature vectors.

\begin{table}[hbt!]
    \caption{Performance on CASIA-B for different partitioning schemes for the Stage 3 module in the GaitPT architecture. We report mean and standard deviation for 5 runs.  For brevity, we only show the average accuracy across all walking angles and variations. There are no significant differences between partitioning schemes (pairwise Welch's t-test $p > 0.05$).}
    \label{tab:stage_3_ablation}
    \begin{center}
    \resizebox{0.85\linewidth}{!}{ 
        \begin{tabular}{l | c }
             \textbf{Stage 3 Partitioning} & \textbf{Mean Accuracy} \\
             \midrule
             Head + Upper Body + Legs & 79.71 \small{$\pm$ 1.3} \\
             Head + Left + Right & 78.57 \small{$\pm$ 0.7} \\
             Head + Opposite Limb Pairing & 79.17 \small{$\pm$ 1.0} \\
             All combinations & 78.48 \small{$\pm$ 1.6}  \\
        \end{tabular}
    }
    \end{center}
\end{table}

The first scheme groups all embeddings for the head area, upper body, and lower body into three distinct feature vectors. The second scheme also computes the head area vector, while combining all joints on the left side of the body into one embedding and the corresponding joints on the right side into another. In the third scheme, we merge opposite limbs, more specifically the left arm with the right leg and the right arm with the left leg, as typically the opposite limbs coordinate with each other in human walking. The final partitioning scheme forces the model to learn interactions between all the combinations described above. Table \ref{tab:stage_3_ablation} showcases mean accuracy across 5 runs on the different partitioning schemes that we explored. We found that there are no significant differences (pairwise Welch's t-test $p > 0.05$) between the variants. We chose to use the first partitioning scheme (Head + Upper Body + Legs) in our experiments to reduce computational time when compared to the "all combinations" partitioning scheme.

\subsection{Pose Estimator Effect on Gait Performance}

\begin{table}[hbt!]
    \caption{Results of the GaitPT trained on skeletal data obtained with different pose estimation models from CASIA-B footage. We report the mean and standard deviation computed across 3 runs. For brevity, we only show the average accuracy across all angles for each walking variation. The final downstream gait recognition accuracy is highly correlated (Pearson's r = 0.919) with the upstream mAP performance of the pretrained pose estimator on the COCO benchmark \cite{lin2014microsoft}.}
    \label{tab:data_quality}
    \begin{center}
    \resizebox{\linewidth}{!}{
    \begin{tabular}{ l | c | c | c }
         \textbf{Pose Estimation Model} & \textbf{COCO mAP} & \textbf{Gait Variation} &  \textbf{Accuracy} \\
        \hline
        \multirow{ 3}{*}{\textbf{OpenPose \cite{OpenPose:Cao}}} &  \multirow{ 3}{*}{64.2} 
          & NM & 70.8 \small{$\pm$ 1.0} \\
        & & BG & 59.1 \small{$\pm$ 0.5} \\
        & & CL & 49.4 \small{$\pm$ 2.1} \\
        \midrule
        
        \multirow{ 3}{*}{\textbf{AlphaPose \cite{alphapose}}} &  \multirow{ 3}{*}{72.3} 
          & NM & 72.7 \small{$\pm$ 2.0} \\
        & & BG & 60.7 \small{$\pm$ 1.7} \\
        & & CL & 55.3 \small{$\pm$ 3.2} \\
        \midrule
        
        \multirow{ 3}{*}{\textbf{YOLOv3 \cite{redmon2018yolov3} + HRNet \cite{sun2019deep}}} &  \multirow{ 3}{*}{77} 
          & NM & 90.0 \small{$\pm$ 1.8} \\
        & & BG & 76.6 \small{$\pm$ 3.0}\\
        & & CL & 71.8 \small{$\pm$ 1.7}\\
        \midrule

        \multirow{ 3}{*}{\textbf{YOLOv7\cite{wang2022yolov7} + ViTPose  \cite{xu2022vitpose}}} &  \multirow{ 3}{*}{79.8} 
          & NM & \textbf{92.3} \small{$\pm$ 0.7} \\
        & & BG & \textbf{79.7} \small{$\pm$ 2.5} \\
        & & CL & \textbf{75.8} \small{$\pm$ 2.0} \\
    \end{tabular}
    }
    \end{center}
\end{table}

Skeleton-based gait recognition is heavily dependent on the performance of the underlying pretrained pose estimation model. Analysis of the patterns of walking requires precise estimation of joint positions across time, in both training and inference. However, to our knowledge, the performance of the pose estimation model and its corresponding quality of skeletons have not been analysed in regard to the gait recognition accuracy. Currently, there is no consensus on the usage of a pose estimator. For instance, GaitGraph \cite{teepe2021gaitgraph} uses HRNet \cite{sun2019deep}, GaitFormer \cite{cosma2022learning} uses AlphaPose \cite{alphapose}. Some datasets also do not release raw RGB videos \cite{cosma2022learning,GREW:zhu,ou-mvlp:Takemura}, and instead release only 2D poses which can rapidly become obsolete as the pose estimation state-of-the-art performance increases. We further present an analysis on the effect the underlying pose estimator has on downstream gait recognition performance. Consequently, we obtained skeletons from all the RGB videos in the CASIA-B dataset, using multiple pose estimation models. 

For this analysis we selected 4 popular pose estimation architectures: \textbf{OpenPose} \cite{OpenPose:Cao}, \textbf{AlphaPose} \cite{alphapose}, \textbf{HRNet} \cite{sun2019deep} and \textbf{ViTPose} \cite{xu2022vitpose}, and trained the GaitPT architecture on the obtained data from each model. We selected these models based on their performance levels on the COCO pose estimation benchmark \cite{lin2014microsoft}, ranging from moderate results (OpenPose) to state-of-the-art performance (VitPose). More recent versions of pose estimation models typically rely on a bounding box of the individuals in the image to make predictions. By using more accurate bounding boxes, we can achieve more precise joint coordinate predictions. For this reason, we opted to utilize \textbf{YOLOv3} \cite{redmon2018yolov3} to obtain inputs for the HRNet model, similarly to Teepe et al. \cite{teepe2021gaitgraph}, and \textbf{YOLOv7} \cite{wang2022yolov7} for the ViTPose architecture. 

Table \ref{tab:data_quality} showcases our results. Across all walking scenarios, the results indicate that higher quality data leads to better recognition accuracy. Remarkably, the accuracy gap between the most inaccurate data obtained with OpenPose and the most reliable data from ViTPose is over 20\% in the normal walking (NM) and the carrying bag (BG) scenarios and 25\% for the clothing (CL) scenario which is regarded as the most difficult. This significant disparity highlights the importance of using a pose estimation model that minimizes data noise. The final gait recognition performance of the models is highly correlated (Pearson's r = 0.919) with the performance of the underlying pose estimation architectures on keypoint detection benchmarks such as COCO \cite{lin2014microsoft} and MPII \cite{andriluka14cvpr}.

\section{Limitations and Societal Impact}
This work has several limitations. GaitPT is trained and evaluated on three datasets that include only a subset of all walking variations present in the real world. Good performance on walking variations captured in laboratory conditions or certain public places does not imply that GaitPT generalizes to the wide range of walking manners across the general population. The training data does not reflect the real-world diversity of individuals, as CASIA-B, GREW, and Gait3D mostly contain people of Asian descent. Regarding potential negative societal impact, gait-based person identification might be used for surveillance and behaviour monitoring. Developments in gait recognition might enable the identification of individuals without their consent. Nevertheless, the development of gait-processing models might aid in the detection of pathological or abnormal gait \cite{gait-survey}.

\section{Conclusions}
In this paper we propose GaitPT, a transformer model designed to process skeleton sequences for gait-based person identification. Compared to other works in the area of gait recognition \cite{teepe2021gaitgraph,cosma2022learning}, GaitPT uses spatial and temporal attention to process the skeleton sequence in a hierarchical manner, guided by the anatomical configuration of the human skeleton. GaitPT is able to process both micro and macro walking movements, which are crucial for fine-grained gait recognition \cite{lin2022gaitgl}. 

We evaluate our architecture on three datasets, corresponding to two scenarios: a controlled laboratory walking scenario on CASIA-B \cite{CASIA:Yu}, to validate the robustness to walking confounding factors, and in realistic, real-world walking scenarios on GREW \cite{GREW:zhu} and Gait3D \cite{zheng2022gait3d}, to test the model's ability to generalize to real-world scenarios. We obtain state-of-the-art performance on both scenarios by a large margin: on CASIA-B we obtain 82.6\% average accuracy (+6\% increase over previous state-of-the-art \cite{teepe2021gaitgraph}), on GREW we obtain 52.16\% rank-1 accuracy (+5.88\% increase over previous silhouette-based state-of-the-art \cite{gait_set:Chao}) and on Gait3D we obtain 13.19\% average rank-1 accuracy (+6.25\% increase over previous best\cite{cosma2022learning}). 

We conduct ablation studies on the most relevant design decision for GaitPT and prove that each stage in the hierarchical pipeline is necessary to obtain good downstream gait recognition performance. We show that there is a high correlation between the performance of the underlying pose estimation model and the downstream performance of the gait recognition model.


{\small
\bibliographystyle{ieee}
\bibliography{bibliography}
}

\end{document}